\documentclass[11pt,a4paper]{article}

\usepackage[hyperref]{emnlp2020}
\usepackage{times}
\usepackage{latexsym}

\usepackage{microtype}

\usepackage{booktabs}
\usepackage{todonotes}
\usepackage{blindtext}
\usepackage{tabularx}
\usepackage{xspace}
\usepackage{amsmath}
\usepackage{enumitem}
\usepackage{colortbl}
\usepackage[normalem]{ulem}

\usepackage{etoolbox,siunitx}	%
\robustify\bfseries
\robustify\itshape
\robustify\uline
\newcommand{\squishlist}{
 \begin{list}{$\bullet$}
  { \setlength{\itemsep}{0pt}
     \setlength{\parsep}{3pt}
     \setlength{\topsep}{3pt}
     \setlength{\partopsep}{0pt}
     \setlength{\leftmargin}{1.5em}
     \setlength{\labelwidth}{1em}
     \setlength{\labelsep}{0.5em} } }

\newcommand{\squishend}{
  \end{list}  }

\definecolor{tablegray}{rgb}{0.2,0.2,0.2}

\newcommand{\tbldomain}[4] {\xspace#1\textcolor{tablegray}{\scriptsize{(#2k, #3)}};}

\aclfinalcopy %
\def\aclpaperid{1032} %

\PassOptionsToPackage{table}{xcolor}

\title{MultiCQA: Zero-Shot Transfer of Self-Supervised Text Matching Models on a Massive Scale}

\author{Andreas R\"uckl\'e \and Jonas Pfeiffer \and
Iryna Gurevych\\
	Ubiquitous Knowledge Processing Lab (UKP)\\
	Department of Computer Science, Technische Universit\"{a}t Darmstadt\\
	{\url{www.ukp.tu-darmstadt.de}}
}

\date{}

\begin{document}
\maketitle
\begin{abstract}

We study the zero-shot transfer capabilities of text matching models on a massive scale, by self-supervised training on 140 source domains from community question answering forums in English.
We investigate the model %
performances on nine benchmarks of answer selection and question similarity tasks, and show that \textit{all} 140 models transfer surprisingly well, where the large majority of models substantially outperforms common IR baselines. %
We also demonstrate that considering a broad selection of source domains is crucial for obtaining the best  zero-shot transfer performances, which contrasts the standard procedure that merely relies on the largest and most similar domains.
In addition, we extensively study how to %
best combine multiple source domains.
 We propose %
 to incorporate 
 self-supervised with supervised multi-task learning on all available source domains. Our best zero-shot transfer model considerably outperforms \emph{in-domain} BERT and the previous state of the art on six benchmarks. 
Fine-tuning of our model with in-domain data results in additional large gains and achieves the new state of the art on all nine benchmarks.

\end{abstract}

\section{Introduction}

Semantic matching of two text sequences is crucial among a wide range of 
NLP problems, such as question answering~\cite{SemEval-2017:task3,Wang2017} or semantic textual similarity~\cite{cer-etal-2017-semeval}.
Due to the ubiquity of applications, 
it is crucial to study how to obtain re-usable text matching models that transfer well to \emph{unseen} domains or %
tasks. %
Zero-shot transfer of text matching models is
particularly challenging %
in setups of non-factoid answer selection \cite{Cohen2018,Tay2017,Feng2015,Verberne2010} and question similarity \cite{SemEval-2017:task3,Lei2016}. These tasks compare questions and answers, or two potentially related questions %
in community question answering (cQA) forums, FAQ pages, and general collections of text passages. %
In contrast to other text matching tasks in NLP, they compare texts of different lengths---e.g., answers can be long explanations or descriptions---%
and often deal with expert domains.
This makes it difficult to transfer models across domains~\cite{shah-etal-2018-adversarial}, and to apply common approaches such as universal sentence embeddings without further domain or task adaptations~\cite{poerner-schutze-2019-multi}.

Non-factoid answer selection and question similarity are also particularly \emph{promising} to study zero-shot transfer. %
Reasons are that (1)~there exist a large number of domains, and (2)~in-domain training data is often scarce. %
Previous work %
proposed
domain adaptation techniques~\cite{poerner-schutze-2019-multi,shah-etal-2018-adversarial}, training with unlabeled data~\cite{rueckle:EMNLP:2019}, and shallow architectures~\cite{rueckle:AAAI:2019}.
However, %
these approaches result in entirely separate models that are specialized to individual target domains. %
One model that is re-usable and targets
zero-shot transfer in similar settings is the question-answer encoder of \citet{yang2019multilingual}, %
which
has recently been evaluated in cross-domain settings for efficient answer sentence retrieval~\citep{m2020multireqa}.
However, they do not study zero-shot transfer with a \emph{large} number of source domains,
and they do not assess how to best combine %
 them. %

In this work, we address these limitations and are---to the best of our knowledge---the first to study the zero-shot transfer capabilities of re-usable text matching models with a large number of source domains %
in these challenging setups.

In the \textbf{first part}, we %
investigate the zero-shot transfer capabilities of 140 domain-specific text matching models to nine benchmark datasets.
By leveraging self-supervised training signals of question title-body pairs, 
we analyze a large number of models specialized on diverse domains.  
We utilize the training method provided by~\citet{rueckle:EMNLP:2019} and train adapter modules~\cite{rebuffi2017learning, houlsby2019parameter} within BERT~\cite{devlin-etal-2019-bert} for \emph{each} of the 140 English StackExchange forums.
Adapters considerably reduce storage requirements by training only a small number of additional parameters while keeping the pre-trained BERT weights fixed.
In our extensive analysis, %
we %
show that
our approach for zero-shot transfer is extremely 
effective---on six benchmarks \emph{all} 140 models outperform common IR baselines. 
Most importantly, we revisit and analyze the traditional strategy of leveraging large data sets from intuitively similar domains to train models for zero-shot transfer. We establish that \emph{neither} training data size \emph{nor} domain similarity are suitable for predicting the best %
models, stressing the need for more elaborate strategies to identify suitable training tasks. This also demonstrates that considering a broad selection of source domains is crucial, which contrasts the standard practice of merely relying on the most similar or largest ones.

\begin{figure}
\centering
\includegraphics[width=\linewidth]{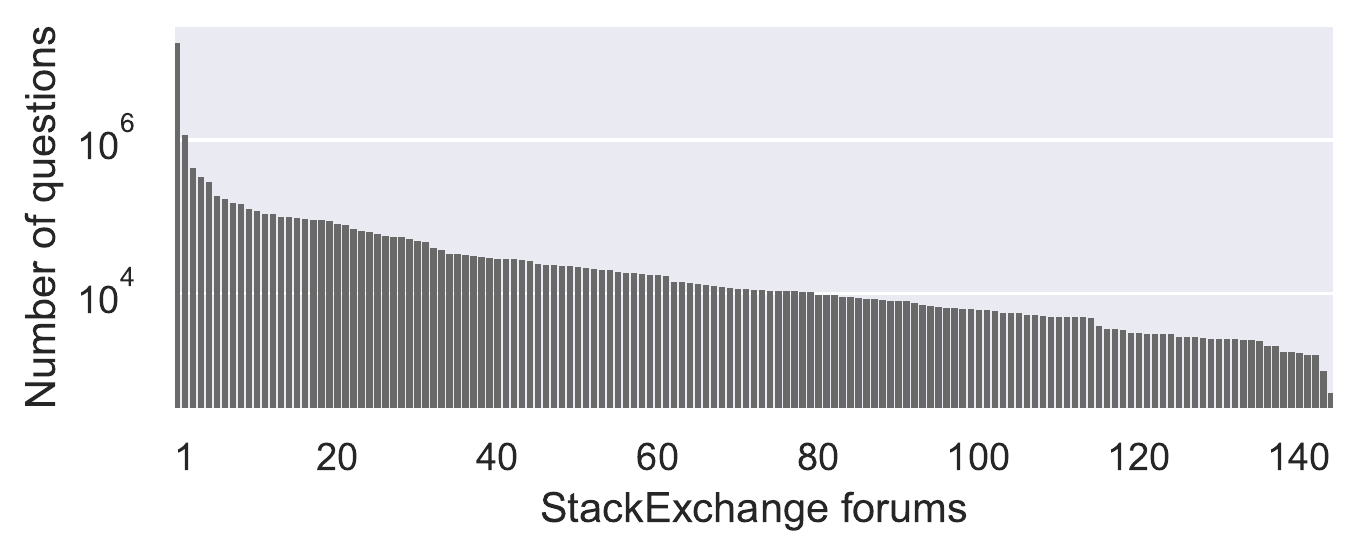}
\caption{Number of questions in StackExchange forums (log scale) that can be used for %
self-supervision. %
}
\label{fig:data:se-stats}
\end{figure}

In the \textbf{second} part of this work, we %
study how to best \emph{combine} multiple source domains with multi-task learning and AdapterFusion~\cite{Pfeiffer2020adapterfusion}.
Our analysis %
reveals that both approaches are not affected by catastrophic interference across training sets.
In particular, our combination of \emph{all} available source domains---despite the large data imbalances, see Figure~\ref{fig:data:se-stats}---is the most effective and outperforms the respective best of 140 single-domain models on six out of nine bechmarks. %
Finally, we combine unlabeled with labeled data for training in a self-supervised and supervised fashion, which considerably improves the zero-shot transfer performances in 16 out of 18 cases.
Our best model substantially outperforms the \emph{in-domain} BERT and RoBERTa~\cite{liu2019roberta} models, as well as the previous state of the art on six benchmarks, which demonstrates its versatility across tasks and domains. %
We also show that our model is an effective initialization for in-domain fine-tuning, which results in large gains %
and achieves state-of-the-art results on all nine benchmarks.

Our source code and the weights of our best multi-task model is publicly available.\footnote{\url{https://github.com/ukplab/emnlp2020-multicqa}} Additionally, all 140 source domain adapters are available at \href{https://AdapterHub.ml}{AdapterHub.ml}~\citep{pfeiffer2020AdapterHub}.

\section{Related Work}

The predominant method %
for text matching tasks such as non-factoid answer selection and question similarity is to train a neural architecture on a large quantity of labeled in-domain data. 
This includes CNN and LSTM models with attention~\cite{Tan2016attention,DosSantos2016attention,Wang2016attention,Rueckle2017:IWCS}, compare-aggregate approaches~\cite{Wang2017,rueckle:AAAI:2019}, and, more recently, transformer-based models~\cite{hashemi2020antique,yang2019multilingual,mass2019study}.
Fine-tuning of large pre-trained transformers such as BERT~\cite{devlin-etal-2019-bert} and RoBERTa~\cite{liu2019roberta} currently achieves state-of-the-art performances on many related benchmarks%
~\cite{garg2019tanda,mass2019study,rochette2019unsupervised,nogueira2019passage}.

However, realistic scenarios often do not provide enough labeled data for supervised in-domain training.
Thus,
different recent work has focused on improving 
model performances in small data scenarios. 
\citet{shah-etal-2018-adversarial} use adversarial domain adaptation %
for duplicate question detection. %
\citet{poerner-schutze-2019-multi} adapt the combination of different sentence embeddings to individual target domains. %
\citet{rueckle:EMNLP:2019} use weakly supervised training, self-supervised training methods, and question generation. %
Similar %
approaches were also explored in %
ad-hoc retrieval~\cite{10.1145/3366423.3380131,ma2020zero,macavaney2019content}. %
A crucial limitation of these approaches is 
that they result in entirely separate models for each dataset and are thus not re-usable.
In this work, we therefore explore the zero-shot transfer capabilities of models, %
to understand how well they generalize to unseen settings.

Previous work of \citet{yang2019multilingual} investigates this on a smaller scale. They propose
USE-QA, a sentence encoder for comparing questions and answers, and achieve promising zero-shot results in retrieval tasks.
However, it is unclear how this model compares to the zero-shot performances of models trained on several different source domains and how best to combine the data from multiple domains.
Other work addresses the generalization of models over several domains in different settings, e.g., for machine reading comprehension~\citep{talmor-berant-2019-multiqa,fisch-etal-2019-mrqa}. %
More related to our work, \citet{m2020multireqa} propose a new evaluation suite with eight datasets for retrieval-based QA, in which they also study the effectiveness of USE-QA. In contrast to them, our work (1) deals with re-ranking setups and uses cross-encoders, which is different to their bi-encoder scenario for retrieval; (2) we deal with question and answer passages instead of answer sentences;
(3)~we study a large number of 140 source domains and provide important insights on zero-shot transfer performances in relation to domain similarity and data size, and extensively analyze the training of models on many source domains simultaneously. %
\section{Data and Setup}

\subsection{Training Data for 140 Domains}
\label{sec:data:ws}
StackExchange is a network that consists of 172 cQA forums,\footnote{See \url{https://stackexchange.com/sites}. The data from  all forums is publicly available \url{https://archive.org/details/stackexchange}} referred to as domains in the following, each devoted to a particular topic such as programming, traveling, finance, etc.
From those 172 forums, 140 are in English and contain more than 1000 unlabeled questions.

We use data from each of these 140 English forums and train domain-specific models %
for semantic text matching.
This has recently become feasible with self-supervised training methods such as \emph{WS-TB}~\cite{rueckle:EMNLP:2019}, in which the question title is considered as a query to retrieve the question body (the detailed description of the question). %
This requires no labeled training instances %
and thus allows us to scale our experiments to 140 source domains  which we can transfer from.

Formally, %
we train models with
positive instances $x^+$ and negative training instances $x^-$:
\begin{align*}
    x_n^+ &= \left( \text{title}(q_n), \text{body}(q_n) \right)
    \\
    x_n^- &= \left( \text{title}(q_n), \text{body}(q_m) \right)
\end{align*}
in which $q_n \neq q_m$. 
We
randomly sample $q_m$ from the entire corpus. %
For computational reasons, we use a maximum of 100k positive training instances.
This training technique performs well for duplicate question detection and answer selection \cite{rueckle:EMNLP:2019}, and similar methods have been %
used for ad-hoc retrieval%
~\cite{macavaney2019content}.

Our different domains are clearly separated by topic. Because not all domains are equally popular, the training sizes are heavily imbalanced, see Figure~\ref{fig:data:se-stats}.
This allows us to analyse the impact of data size in regard to the transfer performances. %

\begin{table}
\footnotesize
\centering
\begin{tabular}{l|S[table-format=5]S[table-format=3]S[table-format=3]|l}
\toprule
 & \multicolumn{1}{|c}{\bfseries Train} & \multicolumn{1}{c}{\bfseries Dev} & \multicolumn{1}{c}{\bfseries Test} & \multicolumn{1}{|l}{\bfseries Source} \\ 
\midrule
\multicolumn{4}{l}{\textit{Non-Factoid Answer Selection}} \\
\midrule
InsuranceQA & 12889 & 1592 & 1625 & In.-Library \\
WikiPassageQA & 3332 & 417 & 416 & Wikipedia \\
LAS-Apple & 5831 & 765 & 766 & StackEx. \\
LAS-Cooking & 3692 & 791 & 792 & StackEx. \\
LAS-Academia & 2856 & 612 & 612 & StackEx. \\
LAS-Travel & 3572 & 765 & 766 & StackEx. \\
LAS-Aviation & 3035 & 650 & 652 & StackEx. \\
\midrule
\multicolumn{2}{l}{\textit{Question Similarity}} \\
\midrule
SemEval17 & 267 & 50 & 88 & QatarLiving \\
AskUbuntu & 12584 & 189 & 186 & StackEx. \\
\bottomrule
\end{tabular}
\caption{The statistics of the evaluation benchmarks.}
\label{table:transfer-datasets}
\end{table}

\subsection{Evaluation Benchmarks}
\label{sec:data:eval}

We transfer all models 
to 9 benchmark datasets from different domains. %
We categorize them in two broad tasks, non-factoid answer selection and question similarity. See Table~\ref{table:transfer-datasets} for the statistics.

\paragraph{Answer selection (AS).}

The goal is to re-rank a pool of candidate answers $A$ in regard to a question $q$. The questions in all datasets are short and do not contain additional descriptions (question bodies). %
Answers to non-factoid questions are often long texts such as descriptions, explanations, and advice.

\squishlist
    \item \emph{InsuranceQA}~\cite{Feng2015} is a benchmark crawled from an FAQ community,\footnote{\url{https://www.insurancelibrary.com/}} in which licensed insurance practitioners answer user questions. The domain is narrow and only contains questions about insurance topics in the US. We use the recent version 2 of the dataset with  $|A|=500$ candidate answers (retrieved with BM25). Typically, one answer is correct.
    \item \emph{WikiPassageQA}~\cite{Cohen2018} %
    was crowd-sourced from Wikipedia articles and is not restricted to a particular domain (although many questions are about history topics). %
    Candidate answers are passages from a single document, on the basis of which the question was formulated. $|A|=58$ of which 1.6 passages represent correct answers (on average).
    \item \emph{Long Answer Selection (LAS)} datasets~\cite{rueckle:AAAI:2019} were crawled from apple, cooking, academia, travel, and aviation StackExchange forums. For a user question, its accepted answer is considered as correct, and negative candidates were collected by retrieving the accepted answers to similar questions (using a search engine with BM25). $|A|=100$ %
\squishend
We measure mean average precision (MAP) on WikiPassageQA, and accuracy (P@1) otherwise. %

\paragraph{Question similarity (QS).}

The goal is to re-rank a pool of potentially related forum questions $C$ in regard to a query question $q$. 
All questions contain titles and bodies---which we concatenate---and are thus long multi-sentence texts. %
On all question similarity benchmarks we measure MAP.

\squishlist
    \item \emph{SemEval17}~\cite{SemEval-2017:task3} refers to Task 3b of the SemEval 2017 challenge. This question similarity benchmark contains %
    instances crawled from QatarLiving forums\footnote{\url{https://www.qatarliving.com/forum}}. For each question $q$, $|C|=10$ potentially related questions were retrieved with a search engine and manually labeled for relatedness in regard to $q$.
    \item \emph{AskUbuntu}~\cite{Lei2016} is an extension of the dataset by \citet{DosSantos2015}, crawled from the AskUbuntu forum. The train split contains noisy community-labeled duplicate annotations, and the (smaller) dev/test splits were manually annotated for relevance. $|C|=20$
\squishend
\begin{figure*}
\centering
\includegraphics[width=\linewidth]{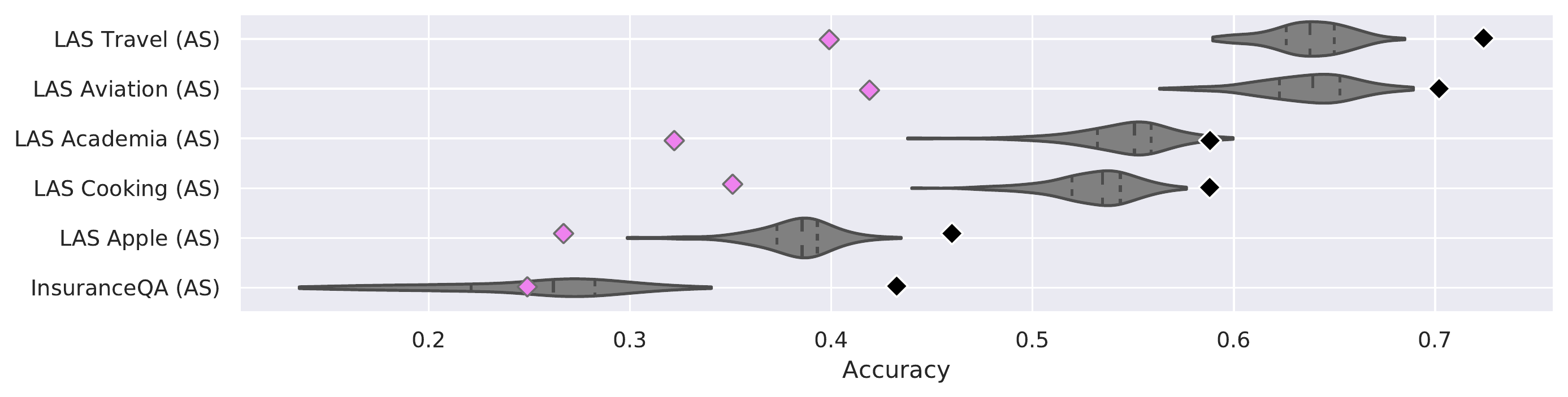}
\includegraphics[width=\linewidth]{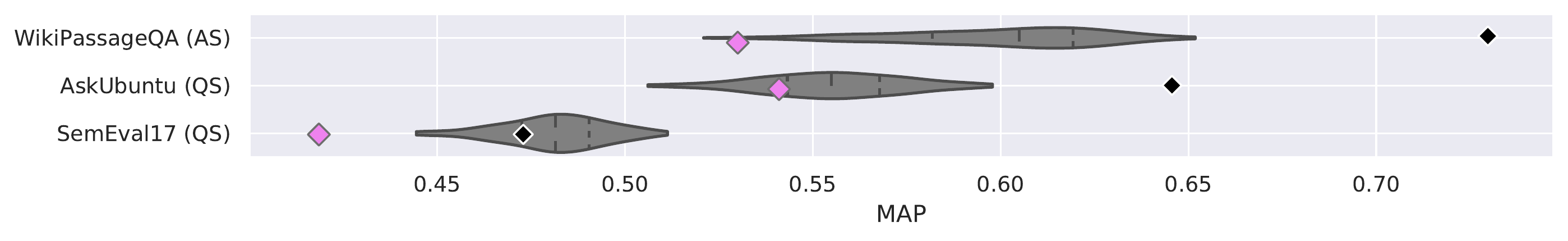}
\caption{Zero-shot transfer performances of all 140 models to the evaluation benchmarks. For benchmarks that contain StackExchange data, we exclude the model from the respective source domain. The violin range visualizes the observed transfer scores, without extension or cut-off for extreme datapoints. Vertical lines show the mean and the quartiles. Diamonds show the performances of IR baselines (violet) and in-domain BERT models (black).}
\label{fig:results:violin}
\end{figure*} 
\subsection{Models and Training}
\label{sec:data:training}

\paragraph{BERT models.}
We
use a pointwise ranking architecture based on pre-trained language models. We concatenate the two input texts (separated with SEP token), and learn a linear classifier on top of the final CLS representation for scoring. We optimize the binary cross-entropy loss.
Similar techniques achieve state-of-the-art results on many related datasets~%
\cite{garg2019tanda,mass2019study,rueckle:EMNLP:2019}.

For our zero-shot transfer experiments from single domains in~\S\ref{sec:st-transfer}, we use BERT base uncased~\cite{devlin-etal-2019-bert}. Later in~\S\ref{sec:mt-transfer}, we additionally investigate BERT large uncased and RoBERTa large~\cite{liu2019roberta}. The hyperparameters for all setups are listed in Appendix~\ref{sec:appendix:hyperparams}.

\paragraph{Training.}
We train our %
models with self-supervision, see~\S\ref{sec:data:ws}. 
To obtain \emph{in-domain models}, we fine-tune BERT with the respective training data of the benchmark datasets of~\S\ref{sec:data:eval}. 
We train the models for 20 epochs with early stopping for in-domain BERT, and without early stopping for zero-shot transfer.
We report the average result over five runs for the in-domain models in AskUbutu and SemEval (due to small evaluation splits) and over two runs for the remaining benchmarks. Following~\citet{mass2019study}, we sample a maximum of 10 negative candidate answers %
for each question in WikiPassageQA (new samples in each epoch). For the LAS datasets we randomly sample 10 negative candidates from the corpus. For InsuranceQA and AskUbuntu, we randomly sample one negative candidate due to their larger training sizes. %

\paragraph{Adapters.}
To reduce the storage requirements, and to efficiently distribute our models to the community, we train adapters~\cite{houlsby2019parameter,rebuffi2017learning} instead of full fine-tuning for our 140 single-domain BERT models. %
Adapters share the parameters of a large pre-trained model---in our case BERT---and introduce a small number of task-specific parameters. With that, adapters transform the intermediate representations in every BERT layer to the training task while keeping the pre-trained model itself unchanged.
We use the recent architecture of~\citet{Pfeiffer2020adapterfusion}, which makes it possible to investigate their adapter combination technique AdapterFusion in \S\ref{sec:mt-transfer}.
In preliminary experiments, we find that using adapters in contrast to full model fine-tuning does not decrease the model performance while drastically reducing the number of parameters (one model is $\sim$5~MB).
\section{Zero-Shot Transfer from 140 Domains}
\label{sec:st-transfer}

In this section, we study the zero-shot transfer performances of all models %
(\S\ref{sec:st-transfer:results}) and investigate whether domain similarity and training data size are suitable for predicting the best models (\S\ref{sec:st-transfer:analysis}).

\subsection{Results}
\label{sec:st-transfer:results}
In Figure~\ref{fig:results:violin}, we show the zero-shot transfer %
to all nine benchmarks. Except for SemEval17, all results are for the dev split.\footnote{SemEval17 does not contain a separate dev split.} Diamonds $\diamond$ show the performance of IR baselines\footnote{TF*IDF for LAS, BM25 for WikiPassageQA and InsuranceQA, and a search engine ranking for SemEval17 (which is the official challenge baseline).} and in-domain BERT.

\paragraph{Zero-shot transfer vs. IR baselines.}
We observe that the wide range of domain-specific models transfer extremely well to all evaluation datasets. For instance, all models largely outperform IR baselines on six benchmarks.
This suggests that learning a \emph{general} similarity function in BERT for our type of data---i.e., short questions and long answers, or pairs of long questions---is important and indeed learned by the models. 
The low variances of the model performances, especially for more general domains such as Travel, Cooking, and SemEval17,
indicate that the domain-specific factors either have a smaller impact, or were already learned during BERT pre-training.
Other work has shown that IR baselines are often hard to beat, e.g., most neural models trained in-domain on WikiPassageQA perform below BM25~\cite{Cohen2018}.
 In contrast, we show that a large number of BERT models from a variety of 140 domains outperform these baselines without requiring any in-domain supervision. %

\paragraph{Zero-shot vs. in-domain models.}
BERT trained in-domain performs the best in most cases.
The difference is larger for expert topics with big training sets (InsuranceQA, AskUbuntu), which shows that our setup provides a challenging test-bed for measuring the generalization capabilities of models. 
However, for target domains with few training instances (see Table~\ref{table:transfer-datasets}), the differences of in-domain BERT to the best zero-shot transfer models are much smaller. Importantly, these setups pose crucial and realistic challenges for text matching approaches%
~\cite{rueckle:AAAI:2019,rueckle:EMNLP:2019}. %
For instance, on 
SemEval17, %
this results in low performances for in-domain BERT. 
In contrast, our best zero-shot transfer model 
achieves a performance of 51.13 MAP---which is 2.13 points better than the best challenge participant in \cite{SemEval-2017:task3}. %

This clearly demonstrates that zero-shot transfer is a suitable alternative for in-domain models,
which also contrasts the large performance degradations often observed with traditional models such as LSTMs \cite{shah-etal-2018-adversarial}. %
Importantly, we find no substantial differences between question similarity and answer selection \emph{tasks}, which are both not explicitly learned during training. %
We thereby take an important step towards overcoming the boundaries between individual tasks and domains. %
\begin{figure*}
\centering
\includegraphics[width=\linewidth]{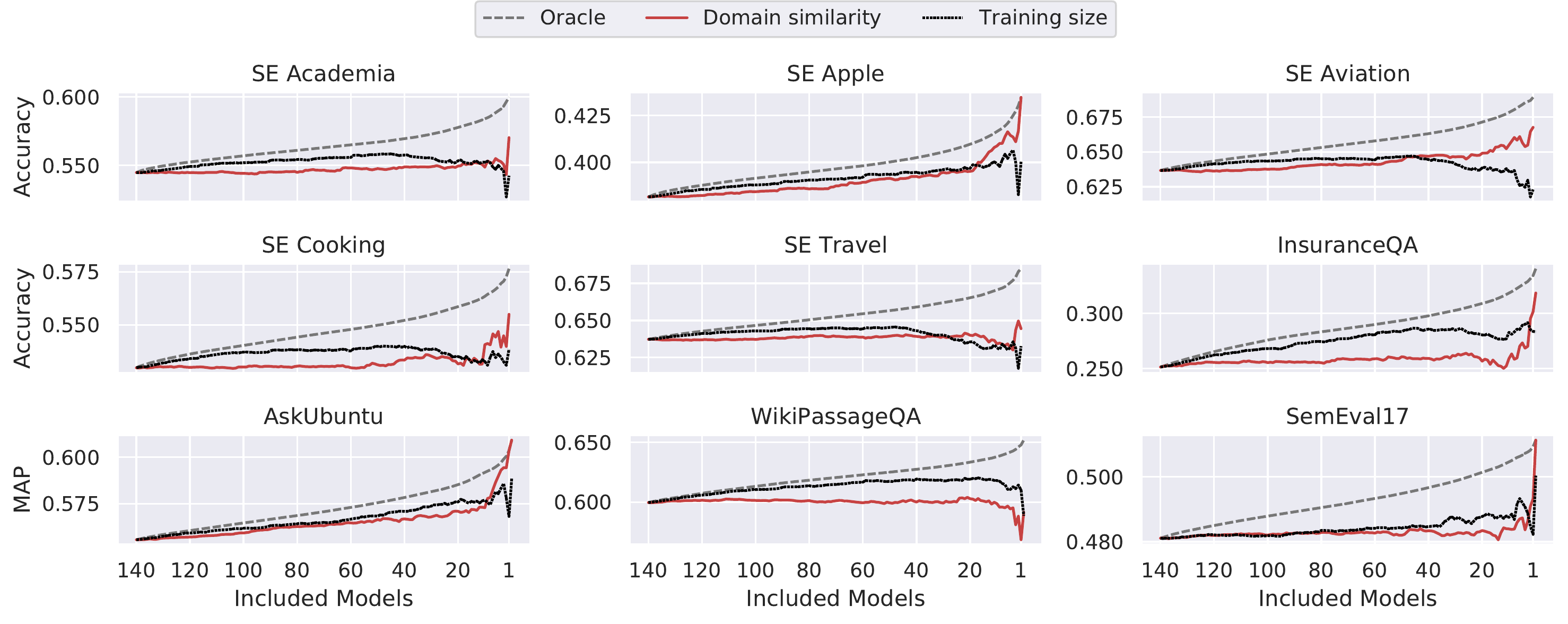}
\caption{The average performance scores (y-axis) of subsets of models (x-axis) selected by domain similarity or training size (scores are averaged over the included models). The oracle always selects the best models. %
}
\label{fig:results:line}
\end{figure*} 

\subsection{Analysis}
\label{sec:st-transfer:analysis}

Due to the large number of 140 domain specific models, each trained on datasets of different sizes, we are able to perform unique analyses regarding the zero-shot transfer performances to target tasks.

Ideally, we would like to identify a small number of models that transfer well to a given dataset, without requiring costly evaluations of all models. 
In the following, we probe the two most commonly used domain selection techniques: (1)~domain similarity and (2)~training size, in regard to the %
transfer performances.
To simulate an optimal selection, we define an \textit{oracle} that always identifies the best models.
We present our findings in Figure~\ref{fig:results:line}.

\paragraph{Domain similarity.}

To measure the domain similarity, we embed the questions of all datasets %
with Sentence-RoBERTa~\cite{reimers-gurevych-2019-sentence}. 
For each dataset, we obtain the mean over all embeddings and %
calculate the domain similarity to other datasets with cosine similarity.

Domain similarity is most effective when selecting models for 
benchmarks of technical domains, e.g., AskUbuntu, LAS-Apple, and LAS-Aviation in Figure~\ref{fig:results:line}.
However, this does not hold true for benchmarks of non-technical domains such as LAS-Travel or WikiPassageQA. 
In those cases, only considering the most similar source domains does not improve the average model performance. %
One reason might be that there do not exist many similar non-technical domains within StackExchange, %
from which models can transfer domain-specific idiosyncrasies. %
However, as we have shown in~\S\ref{sec:st-transfer:results}, such knowledge is not essential, i.e., a large number of models from more distant domains achieve good zero-shot transfer performances.

We provide examples of the best models and the most similar domains for three benchmarks in Table~\ref{table:similar-domains-excerpt} (more are given in Appendix~\ref{sec:appendix:table}).
Many of the best models are from distant domains---e.g., `Ethereum' for WikiPassageQA %
or `SciFi' for LAS-Travel. %
This shows the importance of considering a broad selection of source domains, including ones that are not intuitively close. %
\begin{table*}
\footnotesize
\centering
\begin{tabularx}{\linewidth}{lXp{6.5cm}}
\toprule
\textbf{} & \textbf{Best Models} & \textbf{Most Similar Domains} \\ 
\midrule
\textbf{LAS-Travel} & \tbldomain{scifi}{40}{0.61}{0.68}\tbldomain{money}{18}{0.64}{0.68}\tbldomain{diy}{36}{0.61}{0.67}\tbldomain{space}{7}{0.55}{0.67}\tbldomain{cooking}{14}{0.45}{0.67} & \tbldomain{expatriates}{3}{0.90}{0.64}\tbldomain{law}{9}{0.75}{0.65}\tbldomain{civicrm}{7}{0.75}{0.63}\tbldomain{eosio}{1}{0.73}{0.59}\tbldomain{expressionengine}{7}{0.73}{0.64} \\
\midrule
\textbf{WikiPassageQA} & \tbldomain{\underline{politics}}{6}{0.81}{0.65}\tbldomain{ethereum}{16}{0.61}{0.64}\tbldomain{physics}{158}{0.66}{0.64}\tbldomain{money}{18}{0.57}{0.64}\tbldomain{travel}{24}{0.60}{0.64} & \tbldomain{history}{7}{0.91}{0.59}\tbldomain{literature}{2}{0.84}{0.55}\tbldomain{movies}{13}{0.83}{0.63}\tbldomain{mythology}{1}{0.82}{0.56}\tbldomain{\underline{politics}}{6}{0.81}{0.65} \\
\midrule
\textbf{AskUbuntu} & \tbldomain{\underline{superuser}}{442}{0.81}{0.60}\tbldomain{\underline{apple}}{61}{0.79}{0.59}\tbldomain{blender}{37}{0.54}{0.59}\tbldomain{magento}{70}{0.54}{0.59}\tbldomain{electronics}{85}{0.52}{0.59} & \tbldomain{\underline{superuser}}{442}{0.81}{0.60}\tbldomain{elementaryos}{3}{0.81}{0.58}\tbldomain{\underline{apple}}{61}{0.79}{0.59}\tbldomain{unix}{181}{0.76}{0.59}\tbldomain{serverfault}{288}{0.76}{0.57} \\
\bottomrule
\end{tabularx}
\caption{The best models and the most similar domains for three benchmarks. Parentheses show the training size and the domain similarity (between 0 and 1). \underline{Underlined} domains are in the top-5 of most similar and best models.}
\label{table:similar-domains-excerpt}
\end{table*}
 
\paragraph{Training size.}
The average performance of our models after removing the smallest domains improves more consistently (see WikiPassageQA and InsuranceQA in Figure~\ref{fig:results:line}). %
This shows, that the training size is more suitable for identifying %
models that achieve low performance scores---e.g., models that are trained on very narrow expert domains.  %
However, the training size alone cannot identify the \emph{best} models for zero-shot transfer. %
It is thus crucial to not limit the scope to the largest datasets at hand when exploring suitable training tasks. %
Importantly, this contrasts the common procedure of %
only including the largest domains for transfer~\cite{shah-etal-2018-adversarial}.

\paragraph{Summary} We have established that neither domain similarity nor training data size are suitable for predicting the best models. %
This shows that elaborate strategies are necessary for automatically identifying the most suitable training sets.
Most importantly, we also demonstrate the importance of considering a broad selection of source domains instead of following the standard practice of merely relying on the most similar or largest domains. %
These insights could also be beneficial for researchers in related areas, e.g., to consider a wider range of domains and source datasets %
prior to domain adaptation. %

\begin{table*}
\centering
\footnotesize
\begin{tabular}{ll|S[table-format=2.1]S[table-format=2.1]S[table-format=2.1]S[table-format=2.1]S[table-format=2.1]S[table-format=2.1]!{\color{black!40}\vrule}S[table-format=2.1]|S[table-format=2.2]S[table-format=2.2]S[table-format=2.2]!{\color{black!40}\vrule}S[table-format=2.2]}
\toprule
 & & \multicolumn{1}{c}{\textbf{Tr}} & \multicolumn{1}{c}{\textbf{Co}} & \multicolumn{1}{c}{\textbf{Ap}} & \multicolumn{1}{c}{\textbf{Ac}} & \multicolumn{1}{c}{\textbf{Av}} &  \multicolumn{1}{c!{\color{black!40}\vrule}}{\textbf{IQA}} & \multicolumn{1}{c|}{$\Sigma$} & \multicolumn{1}{c}{\textbf{AU}} & \multicolumn{1}{c}{\textbf{WPQA}} & \multicolumn{1}{c!{\color{black!40}\vrule}}{\textbf{SemEval}} & \multicolumn{1}{c}{$\Sigma$} \\ 
& & \multicolumn{6}{c!{\color{black!40}\vrule}}{\small{\emph{Accuracy scores}}} & & \multicolumn{3}{c!{\color{black!40}\vrule}}{\small{\emph{MAP scores}}} & \\
\midrule
\multicolumn{2}{l|}{Best models of \S\ref{sec:st-transfer}} & 65.4 & 58.1 & 43.1 & 56.5 & 65.0 & 35.3 & 53.9 & 63.29 & 67.85 & 51.13 & 60.75 \\
\midrule
\multicolumn{3}{l}{\textbf{Self-Supervised Training} %
} \\
\midrule
MT & largest & 64.5 & 56.7 & 40.9 & 56.0 & 65.3 & 32.8 & 52.7 & 63.23 & 66.32 & 50.11 & 59.88  \\
AF & largest & 63.4 & 57.7 & 42.9 & 59.1 & 65.9 & 28.6 & 52.9 & 63.11 & 66.99 & \bfseries 50.88 & 60.32 \\
\arrayrulecolor{black!40}
\midrule
\arrayrulecolor{black}
MT & balanced & 65.0 & 60.0 & 43.2 & 55.7 & 65.1 & \bfseries 38.2 & \bfseries 54.5 & \bfseries 63.40 & \bfseries 68.28 & 48.31 & 59.99 \\
AF & balanced & 62.2 & 58.0 & \bfseries 43.5 & \bfseries 59.4 & 66.2 & 29.8 & 53.1 & 62.93 & 67.51 & 47.95 & 59.46 \\
\arrayrulecolor{black!40}
\midrule
\arrayrulecolor{black}
MT & all & \bfseries 66.1 & \bfseries 60.3 & 43.0 & 57.0 & \bfseries 66.4 & 31.5 & 54.0 & 63.32 & 68.20 & 49.81 & \bfseries 60.44 \\
\midrule
\multicolumn{8}{l}{\textbf{Extended Data}} \\
\midrule
MT & balanced & 67.8 & 60.9 & 46.5 & 58.9 & \bfseries 69.1 & \bfseries 34.9 & 56.3 & \bfseries 65.07 & \bfseries 67.70 & 48.59 & 60.45 \\
MT & all &\bfseries 72.4 & \bfseries 63.1 & \bfseries 45.8 & \bfseries 61.1 & 68.0 & 34.7 & \bfseries 57.5 & 64.12 & 66.82 & \bfseries 52.25 & \bfseries 61.06 \\
\bottomrule
\end{tabular}
\caption{Results of MT and AF with different sets of source domains for MultiCQA\textsuperscript{B}. The first five columns are LAS-Travel, Cooking, Apple, Academia, and Aviation. AU is AskUbuntu, IQA is InsuranceQA, WPQA is WikiPassageQA. 
$\Sigma$ shows the average performance of benchmarks that use the same performance measure.
}
\label{table:results-mt-af}
\end{table*}
 
\section{Zero-Shot Transfer from Combinations of Multiple Domains}
\label{sec:mt-transfer}

We now
investigate how to best combine \emph{multiple} source domains for zero-shot transfer. We denote our models as %
\textbf{MultiCQA}. %

\subsection{Setup}
\paragraph{Combination methods.}
We use (1)~multi-task learning and share all model layers across the domains. 
In each minibatch, we sample instances from a single source domain, which we select with a round-robin schedule. Models trained in this manner are denoted as \textbf{MT}.

In addition, we (2)~combine knowledge from our domain adapters (\S\ref{sec:st-transfer}) with AdapterFusion~\cite[\text{\textbf{AF}};][]{Pfeiffer2020adapterfusion}. This learns a weighted combination of multiple (fixed) adapters in each BERT layer and is typically trained on the target task. We adapt this approach to our zero-shot setup and train it with multi-task learning as above.\footnote{We use AdapterFusion without value matrix to avoid additional regularization as in \cite{Pfeiffer2020adapterfusion}.}

\paragraph{Data.}

We use the training data of~\S\ref{sec:data:ws} %
and exclude the domains that are used in any of the evaluation datasets\footnote{AskUbuntu, aviation, travel, cooking, academia, apple.}.
We %
use
three sets of source domains: (1)~the set of 18 topically \textbf{balanced} domains, consisting of the top-three domains (according to the number of questions asked) from each of the six broad categories as defined by StackExchange\footnote{Technology, culture, life, science, professional, and business. See  Appendix~\ref{sec:appendix:domains} for the list of included domains.}; (2)~the \textbf{largest} 18 domains according to the number of asked questions; (3)~\textbf{all} included 134 domains. 

We additionally study the impact of extending our training data with community-labeled instances from the source domains.
For a positive instance of question title and body, we add %
 positive instances of 
(a)~question title and accepted answer, and (b)~question title and body of a duplicate question.
We name this \textbf{extended data}.

\paragraph{Models.}
If not otherwise noted, we fine-tune BERT base. %
We also experiment with BERT large and RoBERTa large (all uncased). For MultiCQA models this corresponds to MultiCQA\textsuperscript{B}, MultiCQA\textsuperscript{B-lg}, and MultiCQA\textsuperscript{RBa-lg}.
The training procedure, number of runs, and hyperparameters are %
as in \S\ref{sec:data:training}.

We additionally compare our models to the question/answer encoder USE-QA~\cite{yang2019multilingual}, which is a state-of-the-art model for %
retrieving answers in zero-shot transfer setups. The IR baselines are the same as in \S\ref{sec:st-transfer:results}
(TF*IDF for LAS, BM25 for WikiPassageQA and InsuranceQA, and a search engine ranking for SemEval17---the official challenge baseline).

\subsection{Results}
\label{sec:mt-transfer:results}

\begin{table*}[!htbp]
\centering
\footnotesize
\begin{tabular}{l|S[table-format=2.1]S[table-format=2.1]S[table-format=2.1]S[table-format=2.1]S[table-format=2.1]S[table-format=2.1]!{\color{black!40}\vrule}S[table-format=2.1]|S[table-format=2.2]S[table-format=2.2]S[table-format=2.2]!{\color{black!40}\vrule}S[table-format=2.2]}
\toprule
 & \multicolumn{1}{c}{\textbf{Tr}} & \multicolumn{1}{c}{\textbf{Co}} & \multicolumn{1}{c}{\textbf{Ap}} & \multicolumn{1}{c}{\textbf{Ac}} & \multicolumn{1}{c}{\textbf{Av}} &  \multicolumn{1}{c!{\color{black!40}\vrule}}{\textbf{IQA}} & \multicolumn{1}{c|}{$\Sigma$} & \multicolumn{1}{c}{\textbf{AU}} & \multicolumn{1}{c}{\textbf{WPQA}} & \multicolumn{1}{c!{\color{black!40}\vrule}}{\textbf{SemEval}} & \multicolumn{1}{c}{$\Sigma$} \\ 
& \multicolumn{6}{c!{\color{black!40}\vrule}}{\small{\emph{Accuracy scores}}} & & \multicolumn{3}{c!{\color{black!40}\vrule}}{\small{\emph{MAP scores}}} & \\
\midrule
IR Baselines & 39.9 & 35.1 & 26.7 & 32.2 & 41.9 & 24.9 & 33.4 & 54.10 & 53.00 & 41.85 & 49.65 \\
\midrule
\multicolumn{3}{l}{\textbf{Zero-Shot Transfer}} \\
\midrule
USE-QA & 65.3 & 58.5 & 44.6 & 46.2 & 53.1 & 35.1 & 50.4 & \bfseries 67.81 & 53.15 & 52.69 & 57.88  \\
MultiCQA\textsuperscript{B} & 72.4 & 63.1 & 45.8 & 61.1 & 68.0 & 34.7 & 57.5 & 64.12 & 66.82 & 52.25 & 61.06 \\
MultiCQA\textsuperscript{B-lg} & 75.5 & 64.6 & 50.0 & 64.0 & 72.0 & 32.8 & 59.8 & 66.48 & 69.83 & 51.56 & 62.62 \\
MultiCQA\textsuperscript{RBa-lg} & \bfseries 77.8 & \bfseries 72.0 & \bfseries 56.8 & \bfseries 70.4 & \bfseries 76.6 & \bfseries 41.9 & \bfseries 65.9 & 63.29 & \bfseries 73.29 &  \bfseries 52.88 & \bfseries 63.15 \\
\midrule
\multicolumn{2}{l}{\textbf{In-Domain Models}} \\
\midrule
Previous SoTA & 69.5$^\dagger$ & 58.3$^\dagger$ & 47.3$^\dagger$ & 58.7$^\dagger$ & 65.5$^\dagger$ & 49.8$^\ddag$ & 58.2 & 69.13$^\dagger$ & 74.90$^\star$ & 51.56$^\diamond$ & 64.76 \\
BERT & 68.7 & 59.0 & 47.0 & 59.0 & 64.5 & 42.2 & 56.7 & 67.31 & 75.09 & 47.29 & 63.23 \\
BERT-lg & 72.5 & 62.4 & 47.2 & 60.0 & 68.3 & 42.7 & 58.8 & 67.54 & 76.22 & 45.86 & 63.20 \\
RoBERTa-lg & 70.9 & 68.4 & 50.7 & 66.3 & 68.7 & 44.9 & 61.6 & 70.18 & 79.74 & 48.70 & 66.20 \\
MultiCQA\textsuperscript{RBa-lg} & \bfseries 80.5 & \bfseries 76.8 & \bfseries 60.2 & \bfseries 72.1 & \bfseries 81.8 & \bfseries 50.8 & \bfseries 70.3 & \bfseries 72.28 & \bfseries 81.41 & \bfseries 53.61 & \bfseries 69.10 \\
\bottomrule
\end{tabular}
\caption{The results of zero-shot transfer and in-domain models. The first five columns are LAS-Travel, Cooking, Apple, Academia, and Aviation. AU is AskUbuntu, IQA is InsuranceQA, and WPQA is WikiPassageQA. 
$\Sigma$ shows the average performance of benchmarks that use the same performance measure.
$^\dagger$~shows the scores of the best BERT models of~\cite{rueckle:EMNLP:2019}, $\ddagger$~is the MICRON model~\cite{han-etal-2019-micron}, $^\star$~is the BERT model in~\cite{ma2019universal}, and $^\diamond$ is MV-DASE~\cite{poerner-schutze-2019-multi}.
}
\label{table:results}
\end{table*} 

\paragraph{Multiple source domains.}
In Table~\ref{table:results-mt-af}, we show the results of MultiCQA\textsuperscript{B} with MT and AF for the different sets of source domains, and compare this to the respective best single-domain models of~\S\ref{sec:st-transfer}.

We observe that the balanced set of source domains achieves better results than combining domains with the largest training sets, which shows that diversity is more important than size.
Most importantly, MT with data from \emph{all} source domains outperforms the respective best single-domain model in 6 out of 9 benchmarks. %
This demonstrates that common problems of MT---%
catastrophic interference between training sets in particular---do not occur in our setup. This also reveals that combining source domains on a massive scale is possible. %

MT and AF are both 
effective combination methods, with minor %
 differences on most datasets. %
However, MT performs considerably better on InsuranceQA, which is a very narrow expert domain. %
The reason for this is that AF combines \emph{fixed} domain-specific adapters, %
which can lead to reduced performances if all adapters are not related to the target domain. 
AF can also lead to better results, e.g., on LAS-Academia. We include an analysis of AF for these datasets in Appendix~\ref{sec:appendix:af}, where we also visualize the learned fusion weights. Interestingly, we find that the fusion weights do not differ much between the two datasets. However, when we remove a single adapter, we also observe that AF automatically replaces it with another adapter from a similar source domain, indicating that this approach is robust. %

\paragraph{Additional labeled data.}
In Table~\ref{table:results-mt-af}, we also see that extending
 the training data of MT models with additional labeled data from question-answer pairs and question duplicates considerably and consistently improves the performances in 16 of 18 cases. This improves the performance of MT~all on \emph{all nine benchmarks}, which shows that our approach is very effective when combining a large number of smaller domains. %
Due to these consistent improvements, we train all our large MultiCQA models with \emph{MT~all} and the extended data. %

\paragraph{Comparison to in-domain models.} 
In Table~\ref{table:results}, we compare our large MultiCQA models  to the in-domain state of the art. %
We find that the additional capacity of the models and the better initialization with RoBERTa considerably improves the zero-shot transfer performances (on average). 
Our best \emph{zero-shot} MultiCQA\textsuperscript{RBa-lg} model outperforms USE-QA on eight benchmarks, and performs better than the previous \emph{in-domain} state of the art on all LAS datasets and on SemEval17. %

Our MultiCQA models are thus highly effective \emph{and} re-usable across different domains and tasks. %
This clearly demonstrates the effectiveness and feasibility of training suitable models for zero-shot transfer that are widely applicable to different realistic settings.

\paragraph{Further in-domain fine-tuning.} 
Finally, we show that MultiCQA\textsuperscript{RBa-lg} is an effective initialization for in-domain fine-tuning. This leads to large gains %
and achieves state-of-the-art results on all nine benchmarks.

\section{Analysis}
\label{sec:analysis}

We manually inspect 50 instances of InsuranceQA and AskUbuntu for which our zero-shot transfer model MultiCQA\textsuperscript{RBa-lg} selects a wrong answer or an unrelated question.
We find that the texts are always on-topic, i.e., many aspects of the question are included in the selected answers (InsuranceQA) or in the potentially similar questions (AskUbuntu). This includes keywords, phrases (often paraphrased), names, version numbers, etc. %
The most common source of error is that an important aspect of the question appears to be ignored or is (likely) not understood by the model. For instance, many aspects of the question might be mentioned in a potentially similar question on AskUbuntu, but in the wrong context. %
Table~\ref{table:analysis:au1} shows an example of such a case, and we provide more examples and additional details in Appendix~\ref{sec:appendix:analysis}. %
We find that this type of error affects 25 of 50 instances in AskUbuntu, and 10 of 50 instances InsuranceQA.\footnote{In 8/50 cases in AskUbuntu and 30/50 cases in InsuranceQA our model actually selects relevant texts, e.g., correct answers or similar questions (which are not labeled as such).}

Future work could thus achieve further improvements by enhancing the overall understanding of question and answer texts. Current models seemingly match similar keywords or phrases of the questions and answers, often without truly understanding them in context. %

\begin{table}
\footnotesize
\centering
\begin{tabularx}{\linewidth}{X}
\toprule
 \textbf{Query question:} Passing parameters to the \uline{installer} for 14.04? The installer for 14.04 gave me no chance (that I took notice of) to pass parameters [...] \\
\midrule
\textbf{Most similar (MultiCQA\textsuperscript{RBa-lg}):} Which key combination would allow me to pass parameters to kernel? During boot I want to pass some parameters like the runlevel , nomodeset to kernel during the \uline{booting process} [...] \\
\midrule
\textbf{Ground truth:} How can i customize the Ubuntu \uline{installer}? I would like to know how can I customize the Ubuntu installer not customize Ubuntu , I just want to modify the installer [...]
\\
\bottomrule
\end{tabularx}
\caption{A mistake of MultiCQA\textsuperscript{RBa-lg} (zero-shot transfer) on AskUbuntu. The model likely does not understand the intention of the query, which is to change the behavior of the installer (and not merely passing parameters to something).}
\label{table:analysis:au1}
\end{table} 
\section{Conclusion}

We studied the zero-shot transfer of text matching models %
on a massive scale, with 140 different source domains and nine benchmark datasets of non-factoid answer selection and question similarity tasks. 
By investigating such a large number of models, we provided an extensive comparison and fair baselines to combination methods, %
and were able to
extensively analyze a large sample size.

We have shown that (1)~BERT models trained in a self-supervised manner on cQA forum data transfer well to all our benchmarks, even across distant domains; %
(2)~training data size and domain similarity are not suitable for predicting the zero-shot transfer performances,  revealing that a broad selection of source domains is crucial;
(3)~our MultiCQA approach that combines self-supervised and supervised training data across a large set of source domains outperforms many in-domain baselines and achieves state-of-the-art zero-shot performances on six benchmarks; 
(4)  fine-tuning MultiCQA\textsuperscript{RBa-lg} in-domain further improves the performances and achieves state-of-the-art results on all nine benchmarks. 

We clearly demonstrated the effectiveness and the relevance of zero-shot transfer in many realistic scenarios and believe that our work lays foundations for a wide range of research questions. %
For instance, combining our approach with additional pre-training objectives such as the Inverse Cloze Task~\citep{chang2020pre} could substantially increase the amount of training data for the large quantity of smaller forums. %
Researchers could also use our 140 domain-specific adapters and investigate further combination techniques to make them even more broadly applicable.

\section*{Acknowledgements}

This work was supported by
the German Federal Ministry of Education and Research (BMBF) and
the Hessen State Ministry for Higher Education, Research and the Arts within their joint support of
the National Research Center for Applied Cybersecurity ATHENE;
the German Research Foundation under grant EC 503/1-1 and GU 798/21-1;
the BMBF %
under the promotional reference 01IS17050 (Software Campus);
and
 the LOEWE initiative (Hesse, Germany) within the emergenCITY center. 

\bibliography{main}

\begin{thebibliography}{39}
\expandafter\ifx\csname natexlab\endcsname\relax\def\natexlab#1{#1}\fi

\bibitem[{Cer et~al.(2017)Cer, Diab, Agirre, Lopez-Gazpio, and
  Specia}]{cer-etal-2017-semeval}
Daniel Cer, Mona Diab, Eneko Agirre, I{\~n}igo Lopez-Gazpio, and Lucia Specia.
  2017.
\newblock \href {https://doi.org/10.18653/v1/S17-2001} {{S}em{E}val-2017 task
  1: Semantic textual similarity multilingual and crosslingual focused
  evaluation}.
\newblock In \emph{Proceedings of the 11th International Workshop on Semantic
  Evaluation ({S}em{E}val-2017)}, pages 1--14, Vancouver, Canada. Association
  for Computational Linguistics.

\bibitem[{Chang et~al.(2020)Chang, Yu, Chang, Yang, and Kumar}]{chang2020pre}
Wei-Cheng Chang, Felix~X Yu, Yin-Wen Chang, Yiming Yang, and Sanjiv Kumar.
  2020.
\newblock \href {https://arxiv.org/abs/2002.03932} {Pre-training tasks for
  embedding-based large-scale retrieval}.
\newblock In \emph{International Conference on Learning Representations (ICLR
  2020)}, Addis Ababa, Ethiopia.

\bibitem[{Cohen et~al.(2018)Cohen, Yang, and Croft}]{Cohen2018}
Daniel Cohen, Liu Yang, and W.~Bruce Croft. 2018.
\newblock \href {https://doi.org/10.1145/3209978.3210118} {{WikiPassageQA}: A
  benchmark collection for research on non-factoid answer passage retrieval}.
\newblock In \emph{The 41st International ACM SIGIR Conference on Research and
  Development in Information Retrieval (SIGIR 2018)}, page 1165–1168, Ann
  Arbor, MI, USA. Association for Computing Machinery.

\bibitem[{Devlin et~al.(2019)Devlin, Chang, Lee, and
  Toutanova}]{devlin-etal-2019-bert}
Jacob Devlin, Ming-Wei Chang, Kenton Lee, and Kristina Toutanova. 2019.
\newblock \href {https://doi.org/10.18653/v1/N19-1423} {{BERT}: Pre-training of
  deep bidirectional transformers for language understanding}.
\newblock In \emph{Proceedings of the 2019 Conference of the North {A}merican
  Chapter of the Association for Computational Linguistics: Human Language
  Technologies (NAACL 2019)}, pages 4171--4186, Minneapolis, Minnesota.
  Association for Computational Linguistics.

\bibitem[{Dos~Santos et~al.(2015)Dos~Santos, Barbosa, Bogdanova, and
  Zadrozny}]{DosSantos2015}
Cicero Dos~Santos, Luciano Barbosa, Dasha Bogdanova, and Bianca Zadrozny. 2015.
\newblock \href {https://doi.org/10.3115/v1/P15-2114} {Learning hybrid
  representations to retrieve semantically equivalent questions}.
\newblock In \emph{Proceedings of the 53rd Annual Meeting of the Association
  for Computational Linguistics and the 7th International Joint Conference on
  Natural Language Processing (ACL 2015)}, pages 694--699. Association for
  Computational Linguistics.

\bibitem[{Dos~Santos et~al.(2016)Dos~Santos, Tan, Xiang, and
  Zhou}]{DosSantos2016attention}
Cicero Dos~Santos, Ming Tan, Bing Xiang, and Bowen Zhou. 2016.
\newblock \href {https://arxiv.org/abs/1602.03609} {{Attentive Pooling
  Networks}}.
\newblock \emph{arXiv preprint arXiv:1602.03609}.

\bibitem[{Feng et~al.(2015)Feng, Xiang, Glass, Wang, and Zhou}]{Feng2015}
Minwei Feng, Bing Xiang, Michael~R. Glass, Lidan Wang, and Bowen Zhou. 2015.
\newblock \href {https://doi.org/10.1109/ASRU.2015.7404872} {Applying deep
  learning to answer selection: A study and an open task}.
\newblock In \emph{2015 IEEE Workshop on Automatic Speech Recognition and
  Understanding (ASRU 2015)}, pages 813--820.

\bibitem[{Fisch et~al.(2019)Fisch, Talmor, Jia, Seo, Choi, and
  Chen}]{fisch-etal-2019-mrqa}
Adam Fisch, Alon Talmor, Robin Jia, Minjoon Seo, Eunsol Choi, and Danqi Chen.
  2019.
\newblock \href {https://doi.org/10.18653/v1/D19-5801} {{MRQA} 2019 shared
  task: Evaluating generalization in reading comprehension}.
\newblock In \emph{Proceedings of the 2nd Workshop on Machine Reading for
  Question Answering}, pages 1--13, Hong Kong, China. Association for
  Computational Linguistics.

\bibitem[{Garg et~al.(2020)Garg, Vu, and Moschitti}]{garg2019tanda}
Siddhant Garg, Thuy Vu, and Alessandro Moschitti. 2020.
\newblock \href {https://arxiv.org/abs/1911.04118} {{TANDA}: Transfer and adapt
  pre-trained transformer models for answer sentence selection}.
\newblock In \emph{Proceedings of the 34th AAAI Conference on Artificial
  Intelligence (AAAI 2020)}. Association for the Advancement of Artificial
  Intelligence.

\bibitem[{Guo et~al.(2020)Guo, Yang, Cer, Shen, and Constant}]{m2020multireqa}
Mandy Guo, Yinfei Yang, Daniel Cer, Qinlan Shen, and Noah Constant. 2020.
\newblock \href {https://arxiv.org/abs/2005.02507} {{MultiReQA}: A cross-domain
  evaluation for retrieval question answering models}.
\newblock \emph{arXiv preprint arXiv:2005.02507}.

\bibitem[{Han et~al.(2019)Han, Choi, Park, and Hwang}]{han-etal-2019-micron}
Hojae Han, Seungtaek Choi, Haeju Park, and Seung-won Hwang. 2019.
\newblock \href {https://doi.org/10.18653/v1/D19-1601} {{MICRON}: Multigranular
  interaction for contextualizing {R}epresentati{ON} in non-factoid question
  answering}.
\newblock In \emph{Proceedings of the 2019 Conference on Empirical Methods in
  Natural Language Processing and the 9th International Joint Conference on
  Natural Language Processing (EMNLP 2019)}, pages 5890--5895, Hong Kong,
  China. Association for Computational Linguistics.

\bibitem[{Hashemi et~al.(2020)Hashemi, Aliannejadi, Zamani, and
  Croft}]{hashemi2020antique}
Helia Hashemi, Mohammad Aliannejadi, Hamed Zamani, and W~Bruce Croft. 2020.
\newblock \href {https://arxiv.org/abs/1905.08957} {{ANTIQUE}: A non-factoid
  question answering benchmark}.
\newblock In \emph{European Conference on Information Retrieval (ECIR 2020)},
  pages 166--173, Lisbon, Portugal. Springer.

\bibitem[{Houlsby et~al.(2019)Houlsby, Giurgiu, Jastrz{k{e}}bski, Morrone,
  de~Laroussilhe, Gesmundo, Attariyan, and Gelly}]{houlsby2019parameter}
Neil Houlsby, Andrei Giurgiu, Stanis{l}aw Jastrz{k{e}}bski, Bruna Morrone,
  Quentin de~Laroussilhe, Andrea Gesmundo, Mona Attariyan, and Sylvain Gelly.
  2019.
\newblock \href {http://proceedings.mlr.press/v97/houlsby19a.html}
  {Parameter-efficient transfer learning for {NLP}}.
\newblock In \emph{Proceedings of the 36th International Conference on Machine
  Learning (ICML 2019)}, pages 2790--2799, Long Beach, California, {USA}.

\bibitem[{Lei et~al.(2016)Lei, Joshi, Barzilay, Jaakkola, Tymoshenko,
  Moschitti, and M{\`a}rquez}]{Lei2016}
Tao Lei, Hrishikesh Joshi, Regina Barzilay, Tommi Jaakkola, Kateryna
  Tymoshenko, Alessandro Moschitti, and Llu{\'\i}s M{\`a}rquez. 2016.
\newblock \href {https://doi.org/10.18653/v1/N16-1153} {Semi-supervised
  question retrieval with gated convolutions}.
\newblock In \emph{Proceedings of the 2016 Conference of the North {A}merican
  Chapter of the Association for Computational Linguistics: Human Language
  Technologies (NAACL 2016)}, pages 1279--1289, San Diego, California.
  Association for Computational Linguistics.

\bibitem[{Liu et~al.(2019)Liu, Ott, Goyal, Du, Joshi, Chen, Levy, Lewis,
  Zettlemoyer, and Stoyanov}]{liu2019roberta}
Yinhan Liu, Myle Ott, Naman Goyal, Jingfei Du, Mandar Joshi, Danqi Chen, Omer
  Levy, Mike Lewis, Luke Zettlemoyer, and Veselin Stoyanov. 2019.
\newblock \href {https://arxiv.org/abs/1907.11692} {{RoBERTa}: A robustly
  optimized {BERT} pretraining approach}.
\newblock \emph{arXiv preprint arXiv:1907.11692}.

\bibitem[{Ma et~al.(2020)Ma, Korotkov, Yang, Hall, and McDonald}]{ma2020zero}
Ji~Ma, Ivan Korotkov, Yinfei Yang, Keith Hall, and Ryan McDonald. 2020.
\newblock \href {https://arxiv.org/abs/2004.14503} {Zero-shot neural retrieval
  via domain-targeted synthetic query generation}.
\newblock \emph{arXiv preprint arXiv:2004.14503}.

\bibitem[{Ma et~al.(2019)Ma, Xu, Wang, Nallapati, and Xiang}]{ma2019universal}
Xiaofei Ma, Peng Xu, Zhiguo Wang, Ramesh Nallapati, and Bing Xiang. 2019.
\newblock \href {https://arxiv.org/abs/1910.07973} {Universal text
  representation from {BERT}: An empirical study}.
\newblock \emph{arXiv preprint arXiv:1910.07973}.

\bibitem[{MacAvaney et~al.(2019)MacAvaney, Yates, Hui, and
  Frieder}]{macavaney2019content}
Sean MacAvaney, Andrew Yates, Kai Hui, and Ophir Frieder. 2019.
\newblock \href {https://doi.org/10.1145/3331184.3331316} {Content-based weak
  supervision for ad-hoc re-ranking}.
\newblock In \emph{Proceedings of the 42nd International ACM SIGIR Conference
  on Research and Development in Information Retrieval (SIGIR 2019)}, page
  993–996, Paris, France. Association for Computing Machinery.

\bibitem[{Mass et~al.(2019)Mass, Roitman, Erera, Rivlin, Weiner, and
  Konopnicki}]{mass2019study}
Yosi Mass, Haggai Roitman, Shai Erera, Or~Rivlin, Bar Weiner, and David
  Konopnicki. 2019.
\newblock \href {https://arxiv.org/abs/1908.06780} {A study of {BERT} for
  non-factoid question-answering under passage length constraints}.
\newblock \emph{arXiv preprint arXiv:1908.06780}.

\bibitem[{Nakov et~al.(2017)Nakov, Hoogeveen, M{\`a}rquez, Moschitti, Mubarak,
  Baldwin, and Verspoor}]{SemEval-2017:task3}
Preslav Nakov, Doris Hoogeveen, Llu{\'\i}s M{\`a}rquez, Alessandro Moschitti,
  Hamdy Mubarak, Timothy Baldwin, and Karin Verspoor. 2017.
\newblock \href {https://doi.org/10.18653/v1/S17-2003} {{S}em{E}val-2017 task
  3: Community question answering}.
\newblock In \emph{Proceedings of the 11th International Workshop on Semantic
  Evaluation ({S}em{E}val-2017)}, pages 27--48, Vancouver, Canada. Association
  for Computational Linguistics.

\bibitem[{Nogueira and Cho(2019)}]{nogueira2019passage}
Rodrigo Nogueira and Kyunghyun Cho. 2019.
\newblock \href {https://arxiv.org/abs/1901.04085} {Passage re-ranking with
  {BERT}}.
\newblock \emph{arXiv preprint arXiv:1901.04085}.

\bibitem[{Pfeiffer et~al.(2020{\natexlab{a}})Pfeiffer, Kamath,
  R{\"{u}}ckl{\'{e}}, Cho, and Gurevych}]{Pfeiffer2020adapterfusion}
Jonas Pfeiffer, Aishwarya Kamath, Andreas R{\"{u}}ckl{\'{e}}, Kyunghyun Cho,
  and Iryna Gurevych. 2020{\natexlab{a}}.
\newblock \href {https://arxiv.org/abs/2005.00247} {{AdapterFusion}:
  Non-destructive task composition for transfer learning}.
\newblock \emph{arXiv preprint arXiv:2005.00247}.

\bibitem[{Pfeiffer et~al.(2020{\natexlab{b}})Pfeiffer, R\"uckl\'{e}, Poth,
  Kamath, Vuli\'{c}, Ruder, Cho, and Gurevych}]{pfeiffer2020AdapterHub}
Jonas Pfeiffer, Andreas R\"uckl\'{e}, Clifton Poth, Aishwarya Kamath, Ivan
  Vuli\'{c}, Sebastian Ruder, Kyunghyun Cho, and Iryna Gurevych.
  2020{\natexlab{b}}.
\newblock \href {https://arxiv.org/abs/2007.07779} {{AdapterHub}: A framework
  for adapting transformers}.
\newblock In \emph{Proceedings of the 2020 Conference on Empirical Methods in
  Natural Language Processing (EMNLP 2020)}, Online.

\bibitem[{Poerner and Sch{\"u}tze(2019)}]{poerner-schutze-2019-multi}
Nina Poerner and Hinrich Sch{\"u}tze. 2019.
\newblock \href {https://doi.org/10.18653/v1/D19-1173} {Multi-view domain
  adapted sentence embeddings for low-resource unsupervised duplicate question
  detection}.
\newblock In \emph{Proceedings of the 2019 Conference on Empirical Methods in
  Natural Language Processing and the 9th International Joint Conference on
  Natural Language Processing (EMNLP 2019)}, pages 1630--1641, Hong Kong,
  China. Association for Computational Linguistics.

\bibitem[{Rebuffi et~al.(2017)Rebuffi, Bilen, and
  Vedaldi}]{rebuffi2017learning}
Sylvestre{-}Alvise Rebuffi, Hakan Bilen, and Andrea Vedaldi. 2017.
\newblock \href
  {http://papers.nips.cc/paper/6654-learning-multiple-visual-domains-with-residual-adapters}
  {Learning multiple visual domains with residual adapters}.
\newblock In \emph{Advances in Neural Information Processing Systems (NeurIPS
  2017)}, pages 506--516, Long Beach, CA, {USA}.

\bibitem[{Reimers and Gurevych(2019)}]{reimers-gurevych-2019-sentence}
Nils Reimers and Iryna Gurevych. 2019.
\newblock \href {https://doi.org/10.18653/v1/D19-1410} {Sentence-{BERT}:
  Sentence embeddings using {S}iamese {BERT}-networks}.
\newblock In \emph{Proceedings of the 2019 Conference on Empirical Methods in
  Natural Language Processing and the 9th International Joint Conference on
  Natural Language Processing (EMNLP 2019)}, pages 3980--3990, Hong Kong,
  China. Association for Computational Linguistics.

\bibitem[{Rochette et~al.(2019)Rochette, Yaghoobzadeh, and
  Hazen}]{rochette2019unsupervised}
Alexandre Rochette, Yadollah Yaghoobzadeh, and Timothy~J Hazen. 2019.
\newblock \href {https://arxiv.org/abs/1911.02645} {Unsupervised domain
  adaptation of contextual embeddings for low-resource duplicate question
  detection}.
\newblock \emph{arXiv preprint arXiv:1911.02645}.

\bibitem[{R{\"u}ckl{\'e} and Gurevych(2017)}]{Rueckle2017:IWCS}
Andreas R{\"u}ckl{\'e} and Iryna Gurevych. 2017.
\newblock \href {https://www.aclweb.org/anthology/W17-6935} {Representation
  learning for answer selection with {LSTM}-based importance weighting}.
\newblock In \emph{Proceedings of the 12th International Conference on
  Computational Semantics ({IWCS} 2017)}. Association for Computational
  Linguistics.

\bibitem[{R{\"u}ckl{\'e} et~al.(2019{\natexlab{a}})R{\"u}ckl{\'e}, Moosavi, and
  Gurevych}]{rueckle:AAAI:2019}
Andreas R{\"u}ckl{\'e}, Nafise~Sadat Moosavi, and Iryna Gurevych.
  2019{\natexlab{a}}.
\newblock \href {https://doi.org/10.1609/aaai.v33i01.33016932} {{COALA}: A
  neural coverage-based approach for long answer selection with small data.}
\newblock In \emph{Proceedings of the 33rd AAAI Conference on Artificial
  Intelligence (AAAI 2019)}, pages 6932--6939. Association for the Advancement
  of Artificial Intelligence.

\bibitem[{R{\"u}ckl{\'e} et~al.(2019{\natexlab{b}})R{\"u}ckl{\'e}, Moosavi, and
  Gurevych}]{rueckle:EMNLP:2019}
Andreas R{\"u}ckl{\'e}, Nafise~Sadat Moosavi, and Iryna Gurevych.
  2019{\natexlab{b}}.
\newblock \href {https://doi.org/10.18653/v1/D19-1171} {Neural duplicate
  question detection without labeled training data}.
\newblock In \emph{Proceedings of the 2019 Conference on Empirical Methods in
  Natural Language Processing and the 9th International Joint Conference on
  Natural Language Processing (EMNLP 2019)}, pages 1607--1617, Hong Kong,
  China. Association for Computational Linguistics.

\bibitem[{Shah et~al.(2018)Shah, Lei, Moschitti, Romeo, and
  Nakov}]{shah-etal-2018-adversarial}
Darsh Shah, Tao Lei, Alessandro Moschitti, Salvatore Romeo, and Preslav Nakov.
  2018.
\newblock \href {https://doi.org/10.18653/v1/D18-1131} {Adversarial domain
  adaptation for duplicate question detection}.
\newblock In \emph{Proceedings of the 2018 Conference on Empirical Methods in
  Natural Language Processing (EMNLP 2018)}, pages 1056--1063, Brussels,
  Belgium. Association for Computational Linguistics.

\bibitem[{Talmor and Berant(2019)}]{talmor-berant-2019-multiqa}
Alon Talmor and Jonathan Berant. 2019.
\newblock \href {https://doi.org/10.18653/v1/P19-1485} {{M}ulti{QA}: An
  empirical investigation of generalization and transfer in reading
  comprehension}.
\newblock In \emph{Proceedings of the 57th Annual Meeting of the Association
  for Computational Linguistics (ACL 2019)}, pages 4911--4921, Florence, Italy.
  Association for Computational Linguistics.

\bibitem[{Tan et~al.(2016)Tan, dos Santos, Xiang, and Zhou}]{Tan2016attention}
Ming Tan, Cicero dos Santos, Bing Xiang, and Bowen Zhou. 2016.
\newblock \href {https://doi.org/10.18653/v1/P16-1044} {Improved representation
  learning for question answer matching}.
\newblock In \emph{Proceedings of the 54th Annual Meeting of the Association
  for Computational Linguistics (ACL 2016)}, pages 464--473, Berlin, Germany.
  Association for Computational Linguistics.

\bibitem[{Tay et~al.(2017)Tay, Phan, Tuan, and Hui}]{Tay2017}
Yi~Tay, Minh~C. Phan, Luu~Anh Tuan, and Siu~Cheung Hui. 2017.
\newblock \href {https://doi.org/10.1145/3077136.3080790} {Learning to rank
  question answer pairs with holographic dual {LSTM} architecture}.
\newblock In \emph{Proceedings of the 40th International ACM SIGIR Conference
  on Research and Development in Information Retrieval (SIGIR 2017)}, page
  695–704, Shinjuku, Tokyo, Japan. Association for Computing Machinery.

\bibitem[{Verberne et~al.(2010)Verberne, Boves, Oostdijk, and
  Coppen}]{Verberne2010}
Suzan Verberne, Lou Boves, Nelleke Oostdijk, and Peter-Arno Coppen. 2010.
\newblock \href {https://doi.org/10.1162/coli.2010.09-032-R1-08-034} {{What Is
  Not in the Bag of Words for Why-QA?}}
\newblock \emph{Computational Linguistics}, 36(2):229--245.

\bibitem[{Wang et~al.(2016)Wang, Liu, and Zhao}]{Wang2016attention}
Bingning Wang, Kang Liu, and Jun Zhao. 2016.
\newblock \href {https://doi.org/10.18653/v1/P16-1122} {Inner attention based
  recurrent neural networks for answer selection}.
\newblock In \emph{Proceedings of the 54th Annual Meeting of the Association
  for Computational Linguistics (ACL 2016)}, pages 1288--1297, Berlin, Germany.
  Association for Computational Linguistics.

\bibitem[{Wang and Jiang(2017)}]{Wang2017}
Shuohang Wang and Jing Jiang. 2017.
\newblock \href {https://arxiv.org/abs/1611.01747} {{A Compare-Aggregate Model
  for Matching Text Sequences}}.
\newblock \emph{International Conference on Learning Representations (ICLR
  2017)}.

\bibitem[{Yang et~al.(2020)Yang, Cer, Ahmad, Guo, Law, Constant, Abrego, Yuan,
  Tar, Sung, Strope, and Kurzweil}]{yang2019multilingual}
Yinfei Yang, Daniel Cer, Amin Ahmad, Mandy Guo, Jax Law, Noah Constant,
  Gustavo~Hernandez Abrego, Steve Yuan, Chris Tar, Yun-hsuan Sung, Brian
  Strope, and Ray Kurzweil. 2020.
\newblock \href {https://doi.org/10.18653/v1/2020.acl-demos.12} {Multilingual
  universal sentence encoder for semantic retrieval}.
\newblock In \emph{Proceedings of the 58th Annual Meeting of the Association
  for Computational Linguistics: System Demonstrations}, pages 87--94, Online.
  Association for Computational Linguistics.

\bibitem[{Zhang et~al.(2020)Zhang, Xiong, Liu, and
  Liu}]{10.1145/3366423.3380131}
Kaitao Zhang, Chenyan Xiong, Zhenghao Liu, and Zhiyuan Liu. 2020.
\newblock \href {https://doi.org/10.1145/3366423.3380131} {Selective weak
  supervision for neural information retrieval}.
\newblock In \emph{Proceedings of The Web Conference 2020 (WWW 2020)}, page
  474–485, Taipei, Taiwan. Association for Computing Machinery.

\end{thebibliography}
\bibliographystyle{acl_natbib}

\clearpage

\appendix
\section{Appendices}
\label{sec:appendix}

\subsection{Hyperparameters}
\label{sec:appendix:hyperparams}
For computational and memory reasons we limit the maximum sequence length to 300 tokens (instead of the maximum of 512 in BERT) for all our models. Similar sequence lengths are commonly used on the benchmarks that we study~\cite[e.g.,][]{mass2019study,Tan2016attention}.

For all experiments, we use a batch size of 32 and a linear warmup schedule over one epoch. We train all models for 20 epochs with early stopping of in-domain models, and without early stopping for zero-shot transfer.

For full model fine-tuning on SemEval17, we use a learning rate of $5 \times 10^{-5}$, due to the very small size of the data set.
In all other cases with full model fine-tuning, we use learning rates that we optimized on WikiPassageQA and InsuranceQA. For this, we explored the manual selection of learning rates of $0.001$, $0.0001$, and $5 \times 10^{-5}$. The development scores on InsuranceQA are  $43.25$, $40.00$, and $39.25$ (accuracy), respectively. The development scores on WikiPassageQA are $72.69$, $71.93$, $72.26$ (MAP), respectively. We thus chose $0.001$ as a learning rate when fine-tuning BERT (and RoBERTa) models.

For the training of adapters and AdapterFusion, we use the learning rates as recommended in~\cite{Pfeiffer2020adapterfusion}, which are $0.0001$ and $5 \times 10^{-5}$, respectively.

\subsection{Computing Infrastructure}
We used a heterogenous cluster with different types of GPUs for our experiments. Our most demanding experiments with RoBERTa-large were performed with one NVIDIA Tesla V100 GPU and 32GB memory (per experiment). To train the models with a batchsize of 32, we used accumulation of gradients over two smaller mini-batches of size 16. One epoch with all source domains trains for on average 97 minutes. The remaining experiments were split across NVIDIA Tesla V100/P100 GPUs (32GB), and NVIDIA Titan RTX (24GB).

\subsection{AdapterFusion (AF) on LAS-Academia and InsuranceQA}
\label{sec:appendix:af}
AdapterFusion learns a weighted combination of adapter outputs in each BERT layer, which is dependent on the layer input. Similar to \citet{Pfeiffer2020adapterfusion}, we can thus plot the activations of the individual adapters for different benchmarks in order to analyze which source domains are most impactful. Further, this allows us to observe how the activations differ across different benchmarks.

\begin{figure*}
\centering
\includegraphics[width=14cm]{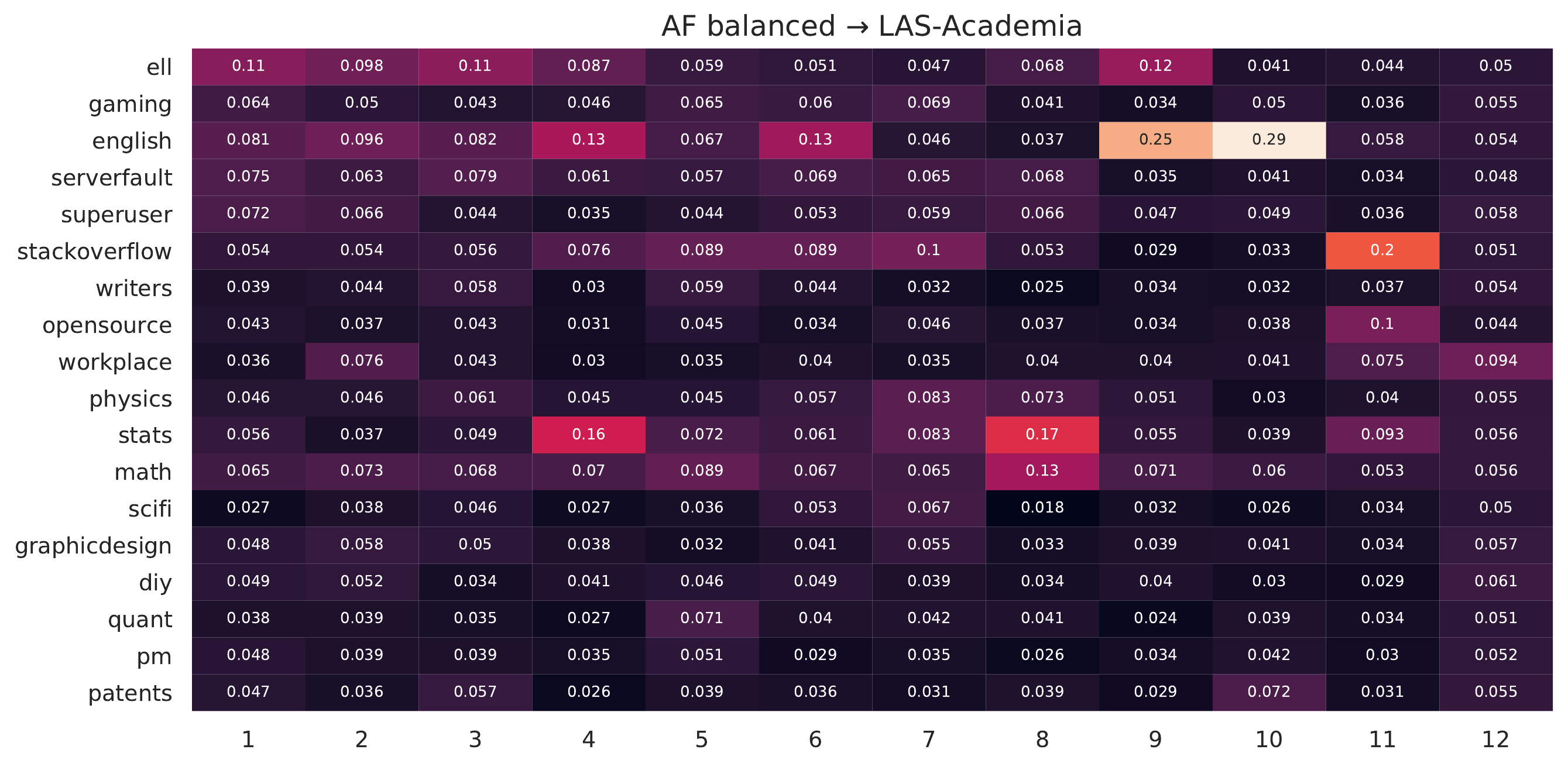}
\caption{Adapter activations in individual BERT layers for AF balanced when transferring to LAS-Academia.}
\label{fig:appendix:af-academia}
\vspace*{\floatsep}%

\includegraphics[width=14cm]{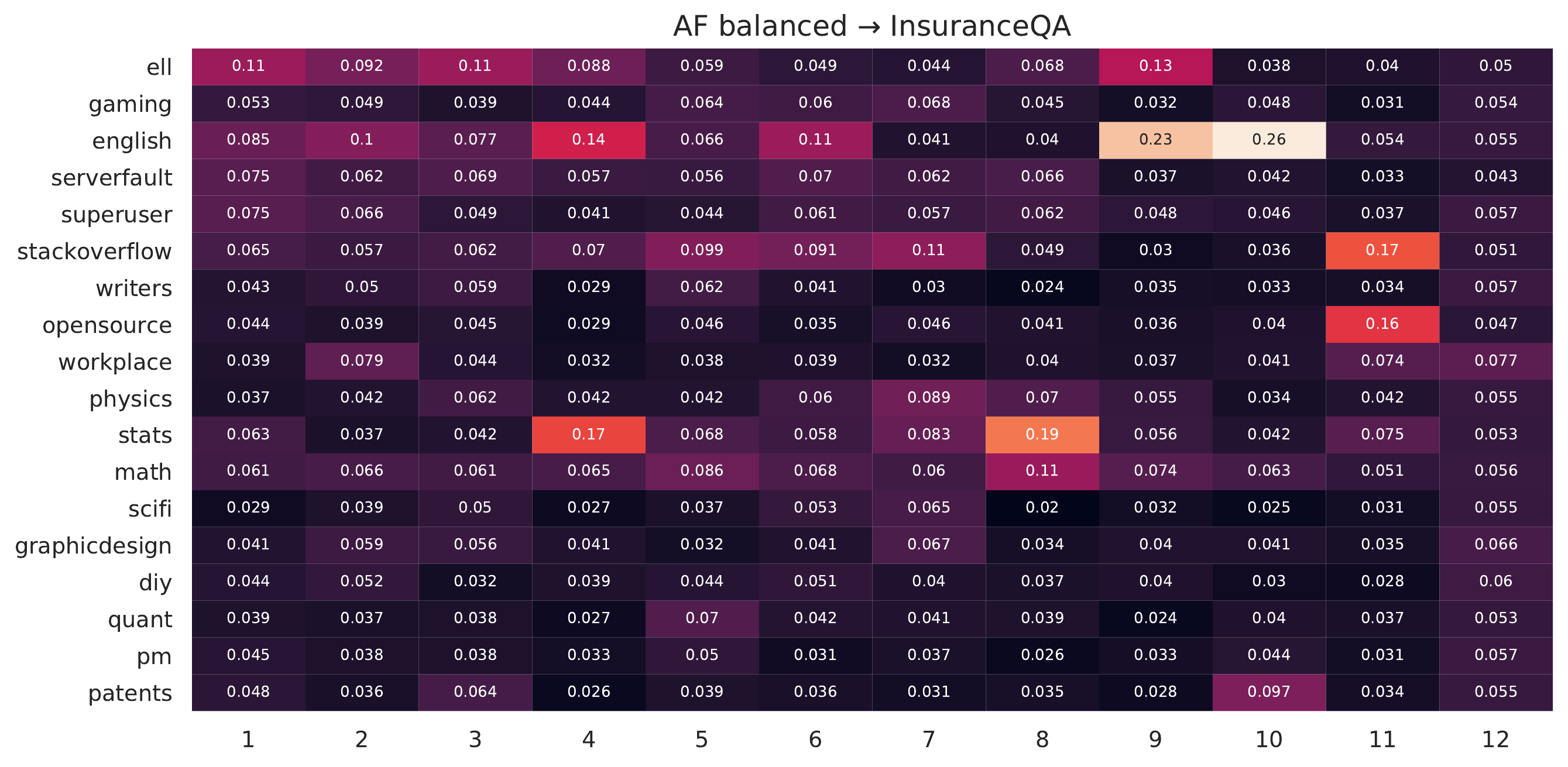}
\caption{Adapter activations in individual BERT layers for AF balanced when transferring to InsuranceQA.}
\label{fig:appendix:af-insurance}
\vspace*{\floatsep}%

\includegraphics[width=14cm]{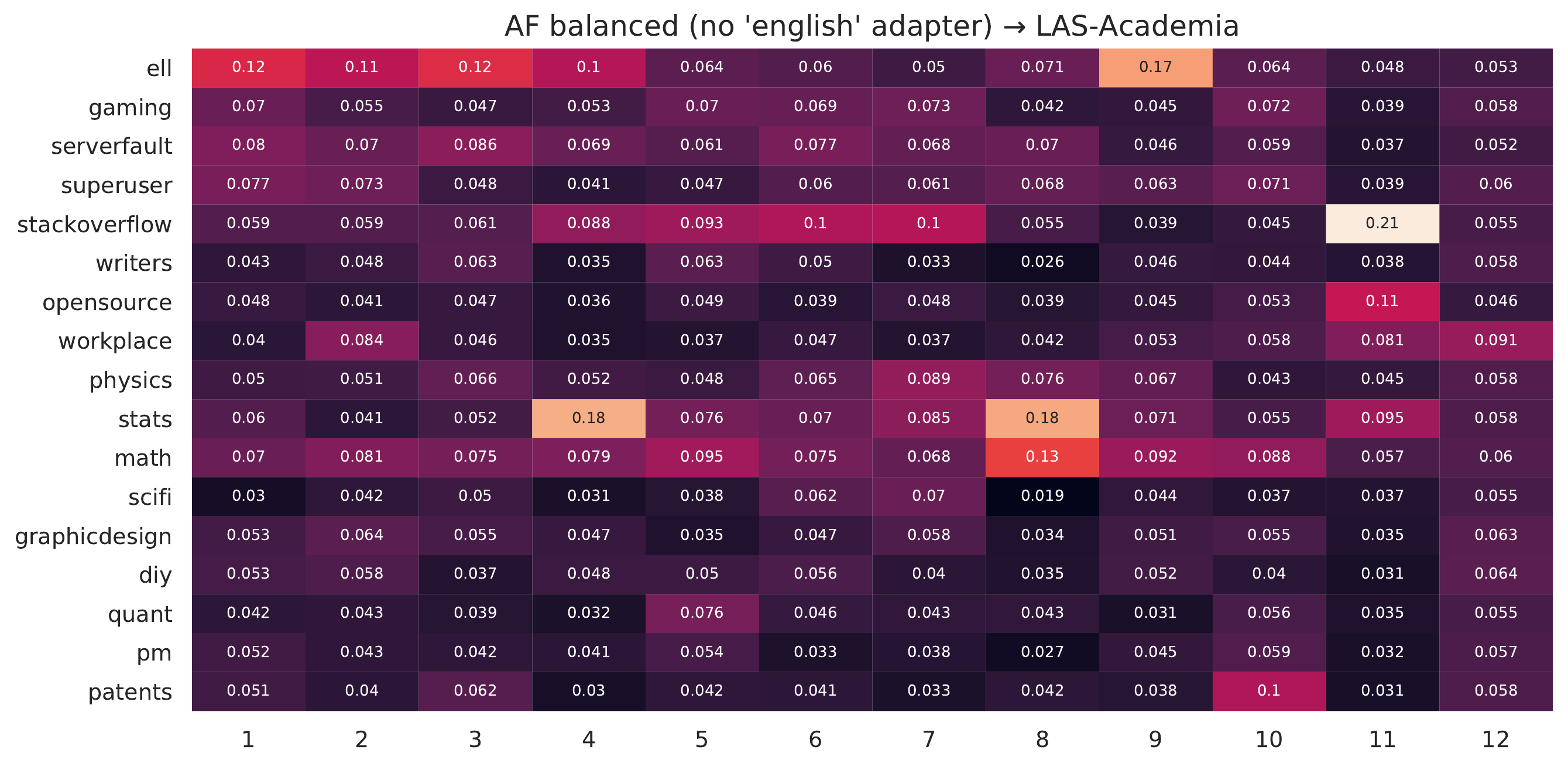}
\caption{Adapter activations in individual BERT layers for AF balanced (excluding the adapter from the `English' domain) when transferring to LAS-Academia.}
\label{fig:appendix:af-academia-2}
\end{figure*}

In Figure~\ref{fig:appendix:af-academia} and in Figure~\ref{fig:appendix:af-insurance} we plot the activatations for \emph{AF balanced} on LAS-Academia and on InsuranceQA, which were the best and worst transfer datasets of this approach, respectively (compared to MT; see \S\ref{sec:mt-transfer:results}).
We find, that the activations are very similar across the two benchmarks, which indicates that our model learns to focus less on the model input. This shows that some adapters are better suited than others for individual BERT layers, e.g., the adapter for the `English' domain dominates layers 9 and 10, and `OpenSource' as well as 'StackExchange' adapters dominate layer 11.

When transferring to the narrow expert domain InsuranceQA, interestingly, the same adapters are activated in BERT layers, with slightly different strengths as compared to LAS-Academia. This means that specific combinations of the same adapters are helpful for a variety of downstream tasks.

To investigate the impact of single most important adapters and how they affect the performance of AF,
we \emph{remove} the adapter of the \emph{English} domain---which has the strongest activations in AF balanced---and plot the result for LAS-Academia in Figure~\ref{fig:appendix:af-academia-2}.
We observe that AF, now increases the activation of the `Ell' (English language learners) adapter (see layer~9).
This shows that AF has learned to utilize particular types of information encoded in adapters that exploit similar attributes, rather than combining a fixed selection of adapters. If, like in this scenario, the adapter is no longer available, AF extracts the information from other, similar adapters. This validates the effectiveness of AF as well as that different kinds of information are stored within the different layers of adapters.

\subsection{Individual Transfer: Best Models and Most Similar Domains}
\label{sec:appendix:table}
We show the best models and most similar domains for all benchmark in Table~\ref{table:similar-domains} (we have provided an excerpt of that in Table~\ref{table:similar-domains-excerpt} of the paper). In particular, we see that the best source domains vary across the different benchmarks. Often, the best models are not from intuitively close domains nor from domains with large training sets.

\begin{table*}[ht]
\footnotesize
\centering
\begin{tabularx}{\linewidth}{lXX}
\toprule
\textbf{} & \textbf{Best} & \textbf{Most similar} \\
\midrule
\textbf{InsuranceQA} & \tbldomain{cooking}{14}{0.38}{0.34}\tbldomain{travel}{24}{0.51}{0.33}\tbldomain{android}{35}{0.39}{0.33}\tbldomain{diy}{36}{0.49}{0.32}\tbldomain{security}{36}{0.51}{0.32} & \tbldomain{money}{18}{0.78}{0.32}\tbldomain{law}{9}{0.69}{0.28}\tbldomain{economics}{5}{0.66}{0.28}\tbldomain{freelancing}{1}{0.66}{0.19}\tbldomain{quant}{8}{0.62}{0.26} \\
\midrule
\textbf{SemEval17} & \tbldomain{\underline{travel}}{24}{0.69}{0.51}\tbldomain{diy}{36}{0.52}{0.51}\tbldomain{gamedev}{31}{0.53}{0.51}\tbldomain{blender}{37}{0.48}{0.51}\tbldomain{gaming}{65}{0.61}{0.51} & \tbldomain{\underline{travel}}{24}{0.69}{0.51}\tbldomain{expats}{3}{0.67}{0.48}\tbldomain{webmasters}{21}{0.64}{0.49}\tbldomain{freelancing}{1}{0.63}{0.47}\tbldomain{workplace}{13}{0.62}{0.47} \\
\midrule
\textbf{WikiPassageQA} & \tbldomain{\underline{politics}}{6}{0.81}{0.65}\tbldomain{ethereum}{16}{0.61}{0.64}\tbldomain{physics}{158}{0.66}{0.64}\tbldomain{money}{18}{0.57}{0.64}\tbldomain{travel}{24}{0.60}{0.64} & \tbldomain{history}{7}{0.91}{0.59}\tbldomain{literature}{2}{0.84}{0.55}\tbldomain{movies}{13}{0.83}{0.63}\tbldomain{mythology}{1}{0.82}{0.56}\tbldomain{\underline{politics}}{6}{0.81}{0.65} \\
\midrule
\textbf{AskUbuntu} & \tbldomain{\underline{superuser}}{442}{0.81}{0.60}\tbldomain{\underline{apple}}{61}{0.79}{0.59}\tbldomain{blender}{37}{0.54}{0.59}\tbldomain{magento}{70}{0.54}{0.59}\tbldomain{electronics}{85}{0.52}{0.59} & \tbldomain{\underline{superuser}}{442}{0.81}{0.60}\tbldomain{elementaryos}{3}{0.81}{0.58}\tbldomain{\underline{apple}}{61}{0.79}{0.59}\tbldomain{unix}{181}{0.76}{0.59}\tbldomain{serverfault}{288}{0.76}{0.57} \\
\midrule
\textbf{LAS-Cooking} & \tbldomain{\underline{gardening}}{8}{0.70}{0.58}\tbldomain{money}{18}{0.43}{0.57}\tbldomain{security}{36}{0.52}{0.57}\tbldomain{academia}{19}{0.38}{0.57}\tbldomain{space}{7}{0.49}{0.56} & \tbldomain{homebrew}{3}{0.82}{0.55}\tbldomain{sustainability}{1}{0.72}{0.52}\tbldomain{health}{4}{0.72}{0.55}\tbldomain{skeptics}{6}{0.71}{0.52}\tbldomain{\underline{gardening}}{8}{0.70}{0.58} \\
\midrule
\textbf{LAS-Apple} & \tbldomain{\underline{superuser}}{442}{0.94}{0.43}\tbldomain{askubuntu}{344}{0.89}{0.43}\tbldomain{\underline{android}}{35}{0.91}{0.42}\tbldomain{\underline{unix}}{181}{0.90}{0.42}\tbldomain{gis}{88}{0.74}{0.41} & \tbldomain{\underline{superuser}}{442}{0.94}{0.43}\tbldomain{windowsphone}{2}{0.93}{0.40}\tbldomain{elementaryos}{3}{0.92}{0.40}\tbldomain{\underline{android}}{35}{0.91}{0.42}\tbldomain{\underline{unix}}{181}{0.90}{0.42} \\
\midrule
\textbf{LAS-Academia} & \tbldomain{scifi}{40}{0.54}{0.60}\tbldomain{money}{18}{0.50}{0.59}\tbldomain{android}{35}{0.51}{0.59}\tbldomain{aviation}{11}{0.52}{0.58}\tbldomain{superuser}{442}{0.58}{0.58} & \tbldomain{writers}{6}{0.78}{0.57}\tbldomain{matheducators}{1}{0.76}{0.52}\tbldomain{workplace}{13}{0.75}{0.57}\tbldomain{softwareengineering}{38}{0.74}{0.56}\tbldomain{pm}{3}{0.74}{0.56} \\
\midrule
\textbf{LAS-Aviation} & \tbldomain{biology}{15}{0.69}{0.69}\tbldomain{diy}{36}{0.68}{0.68}\tbldomain{sports}{3}{0.62}{0.68}\tbldomain{physics}{158}{0.73}{0.68}\tbldomain{rpg}{26}{0.71}{0.68} & \tbldomain{space}{7}{0.81}{0.67}\tbldomain{engineering}{5}{0.76}{0.66}\tbldomain{ham}{1}{0.76}{0.64}\tbldomain{worldbuilding}{14}{0.75}{0.65}\tbldomain{gaming}{65}{0.74}{0.67} \\
\midrule
\textbf{LAS-Travel} & \tbldomain{scifi}{40}{0.61}{0.68}\tbldomain{money}{18}{0.64}{0.68}\tbldomain{diy}{36}{0.61}{0.67}\tbldomain{space}{7}{0.55}{0.67}\tbldomain{cooking}{14}{0.45}{0.67} & \tbldomain{expatriates}{3}{0.90}{0.64}\tbldomain{law}{9}{0.75}{0.65}\tbldomain{civicrm}{7}{0.75}{0.63}\tbldomain{eosio}{1}{0.73}{0.59}\tbldomain{expressionengine}{7}{0.73}{0.64} \\
\bottomrule
\end{tabularx}
\caption{The best models and the most similar domains for all benchmarks. Parentheses show the training size and the domain similarity (between 0 and 1). \underline{Underlined} domains are in the top-5 of most similar and best models.}
\label{table:similar-domains}
\end{table*}

\newpage
\subsection{List of Domains in Combination Experiments}
\label{sec:appendix:domains}
The list of all domains is available on the web: \url{https://stackexchange.com/sites}.
We list the domains used for our two subsets in~\S\ref{sec:mt-transfer} below.

\textbf{Balanced} contains the top-3 domains (according to the number of asked questions) within the six broad categories as defined by StackExchange (technology, culture/recreation, life/arts, science, professional, business).
The included domains are:

ell (English language learners), gaming, english, serverfault, superuser, stackoverflow, writers, opensource, workplace, physics, stats, math, scifi, graphicdesign, diy (do-it-yourself), quant (quantitative finance), pm (project management), patents

\textbf{Largest} contains the top-18 largest domains (according to the number of asked questions).
The included domains are:

stackoverflow, math, superuser, serverfault, latex, unix, physics, statistics, electronics, gis (geographic information systems), english, salesforce, wordpress, magento, sharepoint, gaming, dba (database administrators), drupal

\subsection{Examples of Wrong Predictions on InsuranceQA and AskUbuntu}
\label{sec:appendix:analysis}
We provide additional examples of mistakes made by MultiCQA\textsuperscript{RBa-lg} (zero-shot transfer) on AskUbuntu and InsuranceQA to complement our brief analysis in Section~\ref{sec:analysis}.  %

Table~\ref{table:appendix:analysis:au2} shows an additional example for AskUbuntu. The query question asks for the maximum number of CPUs that can be handled by a kernel. The selected similar question, however, asks for information \emph{where} the kernel gets its information about the available CPUs---not the maximum possible number of CPUs.
Tables~\ref{table:appendix:analysis:inqa1} and \ref{table:appendix:analysis:inqa2} show examples %
of similar problems in InsuranceQA.

\newpage

\begin{table}[!htbp]
\footnotesize
\centering
\begin{tabularx}{\linewidth}{X}
\toprule
 \textbf{Query question:} How many \uline{maximum CPUs} does Ubuntu support by default? I think this is kernel dependent and probably will change over time depending on the kernel a release uses, correct me if wrong I'd like to know~[...] \\
\midrule
\textbf{Most similar (MultiCQA\textsuperscript{RBa-lg})}: Creation of \uline{/proc/stat}. Which function of the kernel creates and writes the information for \uline{/proc/stat}. In this, would like to know \uline{when} kernel gets the CPU information (recognises number of CPUs)~[...] \\
\midrule
\textbf{Ground truth:} Ubuntu Linux 14.04 LTS server edition information need. I was wondering what's the maximum RAM, and \uline{maximum CPUs} does the Ubuntu Linux 14.04 LTS server edition can handle~[...]
\\
\bottomrule
\end{tabularx}
\caption{AskUbuntu example (shortened). This shows that our model mostly focuses on \emph{number of CPUs} and \emph{kernel} information instead of recognizing that the crucial information is the \emph{maximum number}.  We underline important aspects that differ.}
\label{table:appendix:analysis:au2}

\end{table}

\newpage
\begin{table}[!htbp]
\footnotesize
\centering
\begin{tabularx}{\linewidth}{X}
\toprule
 \textbf{Question:} Can I buy a car \uline{without insurance}? \\
\midrule
\textbf{Selected answer (MultiCQA\textsuperscript{RBa-lg}):} You most certainly can get auto insurance \uline{without a car}. if you needed toborrow, test drive, rent, or lease a vehicle for whatever reason you would purchase what is called a drive other car policy.  [...] \\
\midrule
\textbf{Ground truth:} Depending in the state you live in and also if your are financing the car. if you have a loan on the car the financial institution \uline{will require insurance} before you even leave the car lot. if you are buying from a private party they may not require this but in most states you can not even get your license plates with out insurance.
\\
\bottomrule
\end{tabularx}
\caption{InsuranceQA example~1 (shortened). This shows that the model does not interpret the individual keywords within context, i.e., it does not differentiate between \emph{car without insurance} and \emph{insurance without car}. We underline important aspects that differ in the most similar candidate.}
\label{table:appendix:analysis:inqa1}

\vspace*{2\floatsep}

\begin{tabularx}{\linewidth}{X}
\toprule
 \textbf{Query question:} \uline{Why} is state farm life insurance  \uline{so expensive}? \\
\midrule
\textbf{Selected answer (MultiCQA\textsuperscript{RBa-lg}):} State farm  \uline{offers life insurance}, both term and permanent through their captive agents along with property and casualty insurance. However, unlike the latter types of coverage [...] \\
\midrule
\textbf{Ground truth:} Every carrier has their own rates - these are based off a long calculation of actuarial values and mortality tables. Some carriers are more aggressive than others and are willing to take on more risk [...] more conservative carriers feature  \uline{higher rates}. So it's hard to say one carrier is just \uline{very expensive}.
\\
\bottomrule
\end{tabularx}
\caption{InsuranceQA example~2 (shortened). The selected answer describes state farm life insurance, whereas the ground truth explains \emph{why} it can be expensive. We underline important aspects that differ in the most similar candidate.}
\label{table:appendix:analysis:inqa2}

\end{table}
 
\end{document}